\makeatletter
\disable@package@load{breakurl}{}
\makeatother

\documentclass[sn-mathphys-num]{sn-jnl}

\usepackage{placeins}
\usepackage{bbding}
\usepackage{amssymb}
\usepackage{graphicx}%
\usepackage{multirow}%
\usepackage[T1]{fontenc}
\usepackage{stix2}
\usepackage{amsmath,amssymb,amsfonts}%
\usepackage{amsthm}%
\usepackage{pifont} 
\usepackage[title]{appendix}%
\usepackage{xcolor}%
\usepackage{textcomp}%
\usepackage{manyfoot}%
\usepackage{booktabs}%
\usepackage{algorithm}%
\usepackage{algorithmicx}%
\usepackage{algpseudocode}%
\usepackage{listings}%
\usepackage{hyperref}
\usepackage{bm}
\usepackage{natbib}
\usepackage{lineno}

\usepackage[left=2.5cm,top=2.5cm,right=2.5cm,bottom=2.5cm,a4paper]{geometry}

\begin{document}

\title[Article Title]{Skillful Nowcasting of Convective Clouds With a Cascade Diffusion Model}


\author[1]{\fnm{Haoming} \sur{Chen}}\email{hchenda@connect.ust.hk}
\equalcont{These authors contributed equally to this work.}

\author[2]{\fnm{Xiaohui} \sur{Zhong}}\email{x7zhong@gmail.com}
\equalcont{These authors contributed equally to this work.}

\author[1]{\fnm{Qiang} \sur{Zhai}}\email{zhaiq@sicau.edu.cn}

\author[3]{\fnm{Xiaomeng} \sur{Li}}\email{eexmli@ust.hk}

\author[4]{\fnm{Ying Wa} \sur{CHAN}}\email{ywchan@hko.gov.hk}

\author[4]{\fnm{Pak Wai} \sur{CHAN}}\email{pwchan@hko.gov.hk}

\author[1]{\fnm{Yuanyuan} \sur{Huang}}\email{yhuangfh@connect.ust.hk}

\author*[2]{\fnm{Hao} \sur{Li}}\email{lihao\_lh@fudan.edu.cn}

\author*[1,5]{\fnm{Xiaoming} \sur{Shi}}\email{shixm@ust.hk}

\affil[1]{\orgdiv{Division of Environment and Sustainability}, \orgname{The Hong Kong University of Science and Technology}, \orgaddress{\state{Hong Kong}, \country{China}}}

\affil[2]{\orgdiv{Artificial Intelligence Innovation and Incubation Institute}, \orgname{Fudan University}, \orgaddress{\state{Shanghai}, \country{China}}}

\affil[3]{\orgdiv{Department of Electronic and Computer Engineering}, \orgname{The Hong Kong University of Science and Technology}, \orgaddress{\state{Hong Kong}, \country{China}}}

\affil[4]{\orgdiv{Hong Kong Observatory},\orgname{} \orgaddress{\state{Hong Kong}, \country{China}}}

\affil[5]{\orgdiv{Center for Ocean Research in Hong Kong and Macau}, \orgname{The Hong Kong University of Science and Technology}, \orgaddress{\state{Hong Kong}, \country{China}}}

\abstract{Accurate nowcasting of convective clouds from satellite imagery is essential for mitigating the impacts of meteorological disasters, especially in developing countries and remote regions with limited ground-based observations. Recent advances in deep learning have shown promise in video prediction; however, existing models frequently produce blurry results and exhibit reduced accuracy when forecasting physical fields. Here, we introduce SATcast, a diffusion model that leverages a cascade architecture and multimodal inputs for nowcasting cloud fields in satellite imagery.  SATcast incorporates  physical  fields predicted by FuXi, a deep-learning weather model, alongside past satellite observations as conditional inputs to generate high-quality future cloud fields. Through comprehensive evaluation, SATcast outperforms conventional methods on multiple  metrics, demonstrating its superior accuracy and robustness. Ablation studies underscore the importance of its multimodal design and the cascade architecture in achieving reliable predictions. Notably, SATcast maintains predictive skill for up to 24 hours, underscoring its potential for operational nowcasting applications. }

\maketitle{}

\section{Introduction}\label{sec1}

Small-scale deep convection and mesoscale convective systems (MCSs) frequently lead to severe meteorological disasters, such as flooding, wind gusts, and turbulence \cite{chm2024cit, guo2022thunderstorm}.
The United Nation’s "Early Warnings for All" initiative aims to protect global populations from weather hazards by enhancing early warning systems capable of predicting the timing, location, and intensity of such convective events \cite{wmoGuidelinesSatellitebased}.
Satellite observations play an indispensable role in monitoring the formation and evolution of convective systems, offering significantly broader spatial coverage than traditional radar systems.
This advantage is particularly critical in regions with limited radar infrastructure, such as over oceans and in developing countries, where advanced radar networks are sparse or nonexistent \cite{schmit2017closer}.
To address these gaps, satellite-based nowcasting techniques have been developed.
For example, the World Meteorological Organization (WMO) recently issued guidelines for satellite-based nowcasting in Africa, emphasizing its importance in areas with limited observational resources \cite{wmoGuidelinesSatellitebased}.

Traditional methods for satellite and radar-based nowcasting, such as the Lucas-Kanade optical flow algorithm, efficiently track MCS movement but struggle with capturing intensity variations due to assumptions of brightness constancy \cite{baker2004lucas}.
Similarly, systems such as the Short-Term Ensemble Prediction System offer uncertainty estimates but often smooth out small-scale convective details, since it uses stochastic noise to model the evolution of small-scale convection \cite{smith2024evaluating}.
These limitations have catalyzed the adoption of deep learning models, which do not have these assumptions and excel in producing more reliable forecasts \cite{zhang2023skilful, chen2023fuxi, bi2023accurate}.

The evolution of deep learning in nowcasting has yielded promising advances, although certain challenges persist.
Forecast horizons for satellite or radar-based predictions rarely exceed six hours due to cumulative errors, and deep learning models often produce blurred outputs \cite{kumar2020convcast,tran2019computer,ehsani2021nowcasting,wei2024dayudatadrivenmodelgeostationary}.
For instance, Shi et al. \cite{shi2015convolutional} introduced convolutional Long Short Term Memory (ConvLSTM) networks for radar image time series forecasting, demonstrating superior performance in precipitation nowcasting with lead times of up to 2 hours.
More recently, DaYu, which integrates transformer \cite{vaswani2017attention} and residual convolution layers, has extended the skillful forecast lead time for satellite-based nowcasting to 12 hours \cite{wei2024dayudatadrivenmodelgeostationary}.
However, its outputs become overly smooth after 3 hours, particularly in representing typhoon structure and the eye's position (see Figures 3 and 4). Wang et al. \cite{wang2023skillful} used Generative Adversarial Networks (GAN) \cite{goodfellow2020generative} to generate high-quality radar images successfully,achieving skillful forecast up to 2 hours. Kim et al. \cite{KIM2024105529} incorporated European Centre for Medium-Range Weather Forecasts Reanalysis version 5 (ERA5) data into a deep learning model, improving precipitation forecasts within six hours compared to those generated using only radar images inputs. However, the production schedule of ERA5 data is not compatible with the requirements of nowcasting, and they \cite{KIM2024105529} suggest that numerical weather prediction (NWP) data could serve as a viable alternative. Despite this, NWP models also face significant challenges including high computational costs associated with running NWP models, and uncertainties in parameterization schemes for gray-zone convection, which limit their ability to accurately forecast convective activities \cite{trier2014use,2022sxm}.
As a result, weather data-driven nowcasting techniques that operate independently of NWP models offer a more cost-effective and practical solution \cite{chen2023fuxi,bi2023accurate}. 

Physical variables generated by either deep learning or NWP models exhibit significant differences in information content compared to radar and satellite observations, making them multimodal data sources. The integration of multimodal data is a key method for enhancing model performance, and further research is needed on how to effectively process and combine these diverse data types in deep learning models for atmospheric science.
For example, the Contrastive Language-Image Pretraining model employs transformer and convolutional layers to encode text and images, enabling a more flexible classifier that serves as a foundation for subsequent multimodal models \cite{2021Learningclip}.
In atmospheric science, MetNet-3 \cite{Andrychowicz2023DeepLF} employs different ResNet blocks \cite{2016Deep} to process multi-resolution data, thereby extending skillful forecast lead times up to 24 hours. Wang et al. \cite{2024wangdiff} uses a cross-attention module to merge NWP results and radar images, extending precipitation forecasts up to six hours. Therefore, integrating physical variables and satellite imagery into deep learning models is beneficial, but bridging the gap between these data types remains challenging.

In this study, we introduce SATcast for predicting convective clouds observed by the FengYun-4A (FY-4A) geostationary satellite. Diffusion models, which are state‐of‐the‐art in text-to-image and video generation in computer vision \cite{ho2020denoising, dhariwal2021diffusion}, are employed for this purpose.
The FY-4A satellite plays a crucial role in monitoring weather patterns, climate variations, environmental changes, and natural disasters across East Asia \cite{yang2017introducing}.
Our physical‐driven model innovatively incorporates atmospheric physical conditions predicted by FuXi \cite{chen2023fuxi}, which has demonstrated remarkable accuracy in deterministic weather predictions, as conditioning inputs.
Additionally, we employ a cascade structure to mitigate the error accumulation in nowcasting.
This multimodal, cascade-based deep learning nowcasting model achieves remarkable performance in predicting convective clouds with forecast lead times of up to 24 hours, addressing the long-standing issues of blurriness in nowcasting.

\section{Results}\label{sec2}

We train our nowcasting model using four years (2019-2022) of hourly FY-4A satellite data from the infrared channel (channel 12) of the Advanced Geosynchronous Radiation Imager (AGRI).
Channel 12 is selected as it is specifically designed for water vapor detection and mid-troposphere monitoring, making it particularly effective for tracking cloud system development and convective processes \cite{lu2017fy,yan2024study}.
The diffusion model employs a U-Net structure enhanced with ConvNeXt modules \cite{liu2022convnet} and self-attention modules \cite{vaswani2017attention} (Fig.~\ref{fignet}).

The model combines FuXi forecasts of atmospheric dynamic and thermodynamic fields with historical satellite imagery to predict cloud evolution.
Detailed information about the specific FuXi forecast variables used is provided in Supplementary Table 1.
The primary model, SATcast, implements a cascade, autoregressive, and multimodal framework.
SATcast (referred to as SATcast-phase 1) generates four-hour forecasts (T+1 to T+4) of satellite imagery based on 12 hours (T$-$7 to T+4) of FuXi forecasts and eight hours (T$-$7 to T) of historical satellite imagery.
To extend forecasts beyond the initial four-hour period, the model iteratively uses its outputs as inputs for subsequent predictions.
To improve forecast accuracy for longer lead times, a separate model, SATcast-phase 2, which is fine-tuned specifically for lead times of 5 to 8 hours, is employed in a cascaded framework.
This model shares the identical structure as SATcast-phase 1 but uses independent parameters optimized for longer time predictions.
Further details regarding the model design, training process, and evaluation methodology are provided in Section \ref{sec11}.

Furthermore, to assess the contributions of key components in the framework, we conduct ablation studies using three model variants: SATcast-NoC, which removes the cascade and autoregressive framework and generates eight-hour forecasts in a single step; SATcast-NoF, which excludes FuXi forecasts as conditioning inputs; SATcast-NoT, which excludes the cascaded model for longer lead times. For baseline models, we used persistence model, which assumes that convective cloud patterns remain constant across all lead times \cite{lagerquist2021using}, and optical flow, a popular method in practical nowcasting. The results from optical flow are compared with those of SATcast. Model performance is evaluated from multiple perspectives, including statistical metrics, case studies, the clarity of weather system structures.

\subsection{Overall performance of SATcast}
\label{sec22}

Previous studies have employed multiple deep learning models, each tailored to specific lead time intervals, to mitigate the accumulation of prediction errors \cite{chen2023fuxi,bi2023accurate}.
In this study, constrained by computational resources, we train only two models within a cascade framework.
SATcast-phase 1 predicts cloud evolution for lead times from T+1 to T+4, while SATcast-phase 2 utilizes outputs from SATcast-phase 1 to generate predictions for lead times from T+5 to T+8.
For lead times beyond T+8, we evaluate model performance by iteratively applying SATcast-phase 1 to generate satellite image forecasts.

\begin{figure}[!hb]
\centering
\noindent\includegraphics[width=0.85\linewidth]{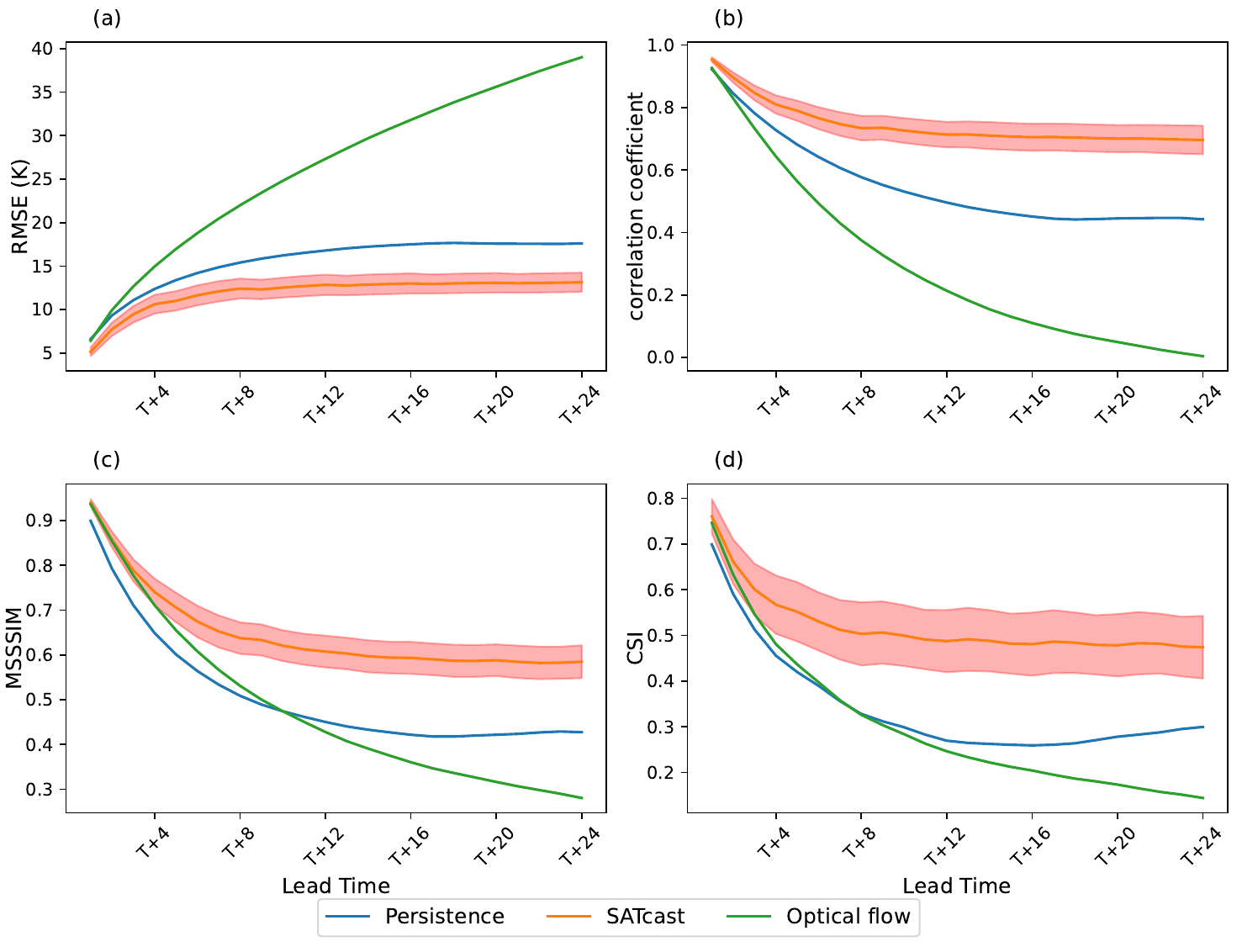}
\caption{Comparison of the RMSE, correlation coefficient, MSSSIM, and CSI spatially averaged over (86$^\circ$ to 150$^\circ$ E in longitude and 1$^\circ$ to 41$^\circ$ N in latitude. The results from persistence model, SATcast, and optical flow. The results using testing data from 2022 September to 2022 December. The threshold is 240\, K in the calculation of CSI. The red shading represents the one standard deviation range of SATcast.} 
\label{fig1}
\end{figure}
\FloatBarrier

Figure \ref{fig1} compares the root mean square error (RMSE), correlation coefficient, multi-scale structure similarity (MSSSIM) \cite{1292216}, and critical success index (CSI) between SATcast, the persistence model, and the optical flow method.
The CSI is calculated using a threshold of 240 K.
Results are spatially averaged over the region of interest (86$^\circ$ to 150$^\circ$ E in longitude and 1$^\circ$ to 41$^\circ$ N in latitude), based on 512 testing sequences, each spanning 32 hours (the first 8 hours serve as input, and the following 24 hours as the target), from September to December 2022.
The red shading in SATcast denotes the one standard deviation calculated across lead time ranges for samples with different initialization times.

\begin{figure}[b]
\centering
\noindent\includegraphics[width=\linewidth]{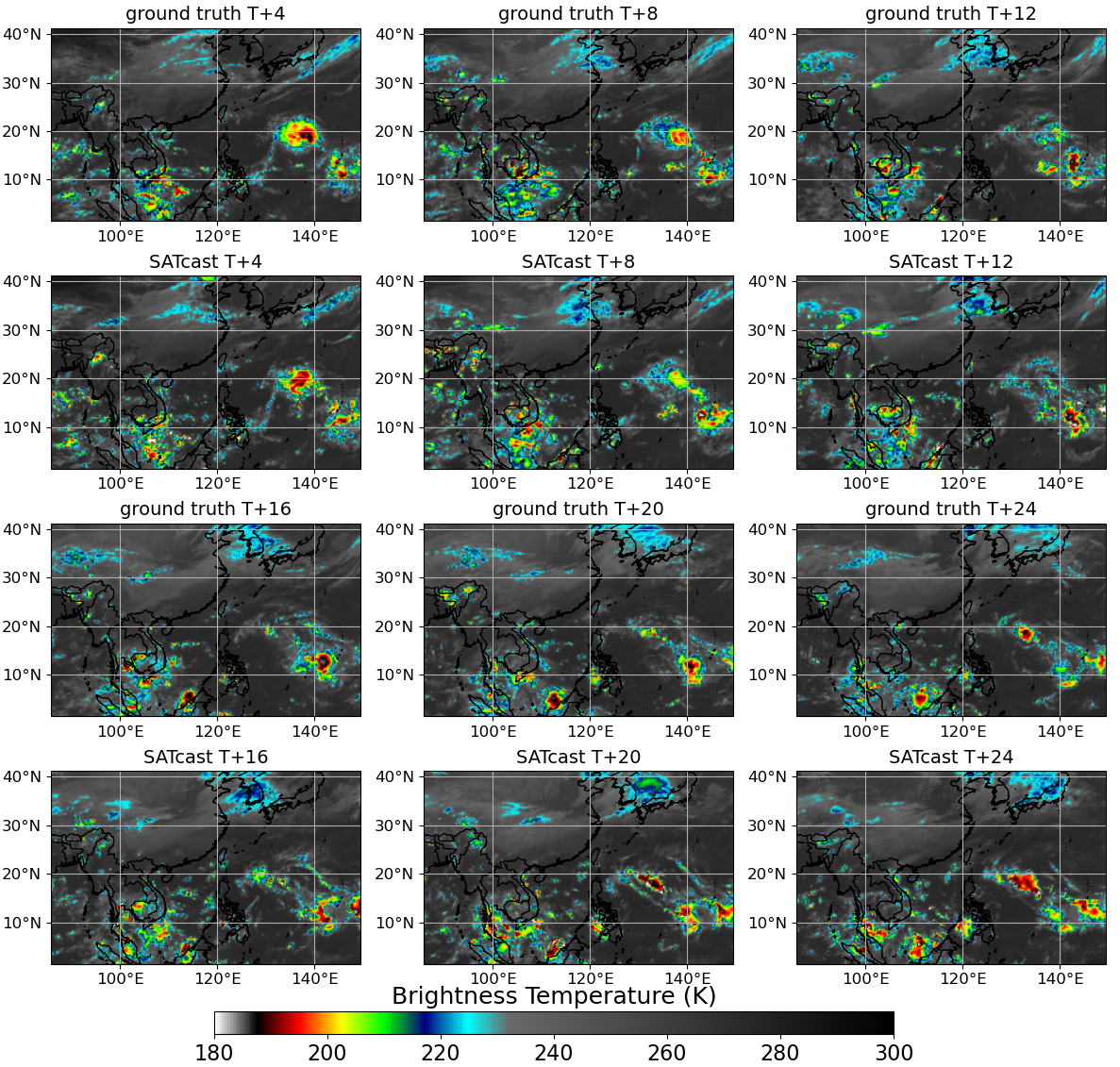}
\caption{Spatial distribution of brightness temperature from observation and SATcast predictions, shown at 4-hour time intervals from T+4 to T+24.} 
\label{fig5}
\end{figure}

As shown in Figure \ref{fig1}, all metrics for SATcast and the persistence model degrade sharply up to T+8 but stabilize after T+12. In contrast, the performance of the optical flow method decreases consistently.
Notably, the persistence model exhibits slight performance improvements between T+20 and T+24, likely due to the influence of the diurnal cycle \cite{wallace1975diurnal}.
Overall, SATcast consistently outperforms both the persistence model and the optical flow method across all metrics.
The correlation coefficient for SATcast stabilizes above 0.7, significantly surpassing the persistence baseline, which drops below 0.5 after T+12.
Similarly, SATcast achieves a CSI of around 0.5, indicating reliable detection of convective clouds, while the persistence model's CSI falls below 0.3.
The MSSSIM for SATcast remains approximately 0.6, suggesting that the quality of predicted cloud fields remains high even as the lead time increases. In contrast, the MSSSIM for the persistence model is about 0.45, due to variations in weather systems.
These results demonstrate that the cascade framework with single-round fine-tuning effectively reduces cumulative errors, resulting in minimal degradation of predictive performance beyond T+8.
The stability of SATcast's metrics beyond T+12 further underscores its reliability for longer-term cloud predictions.

Figure \ref{fig5} demonstrates the SATcast's 24-hour forecasting capability using a typhoon case. The spatial distribution of brightness temperature from the optical flow method becomes very smooth after several time steps.
We focus on SATcast's performance, with the optical flow results presented in Supplementary Figure S12.
SATcast successfully captures the spatiotemporal evolution of a large-scale convection system moving from northern China to Japan.
It also accurately predicts two tropical disturbances east of the Philippines, which latter intensified into Typhoon Sonca and Typhoon Nesat.
The model reproduces the complex intensity fluctuations, showing an initial weakening followed by intensification, due to interactions with cold eddies, within the 24-hour period.
Predicted typhoon intensity and track closely align with observations, as shown in Supplementary Figure S1, with additional cases provided in Supplementary Figures S2--S3.
Notably, SATcast achieves these high-fidelity predictions with a single round of fine-tuning, effectively mitigating forecast error accumulation.
This continuous forecasting capability circumvents prediction discontinuities, suggesting SATcast as a reliable framework for convective cloud forecasting.

In addition, SATcast demonstrates strong generalization capability.
Despite being trained exclusively on channel 12 data, it effectively predicts the brightness temperature on channel 9 by incorporating FuXi data.
Relevant metrics and examples are provided in Supplementary Figures S9-S10.

\begin{figure}[!b]
\centering
\noindent\includegraphics[width=0.85\linewidth]{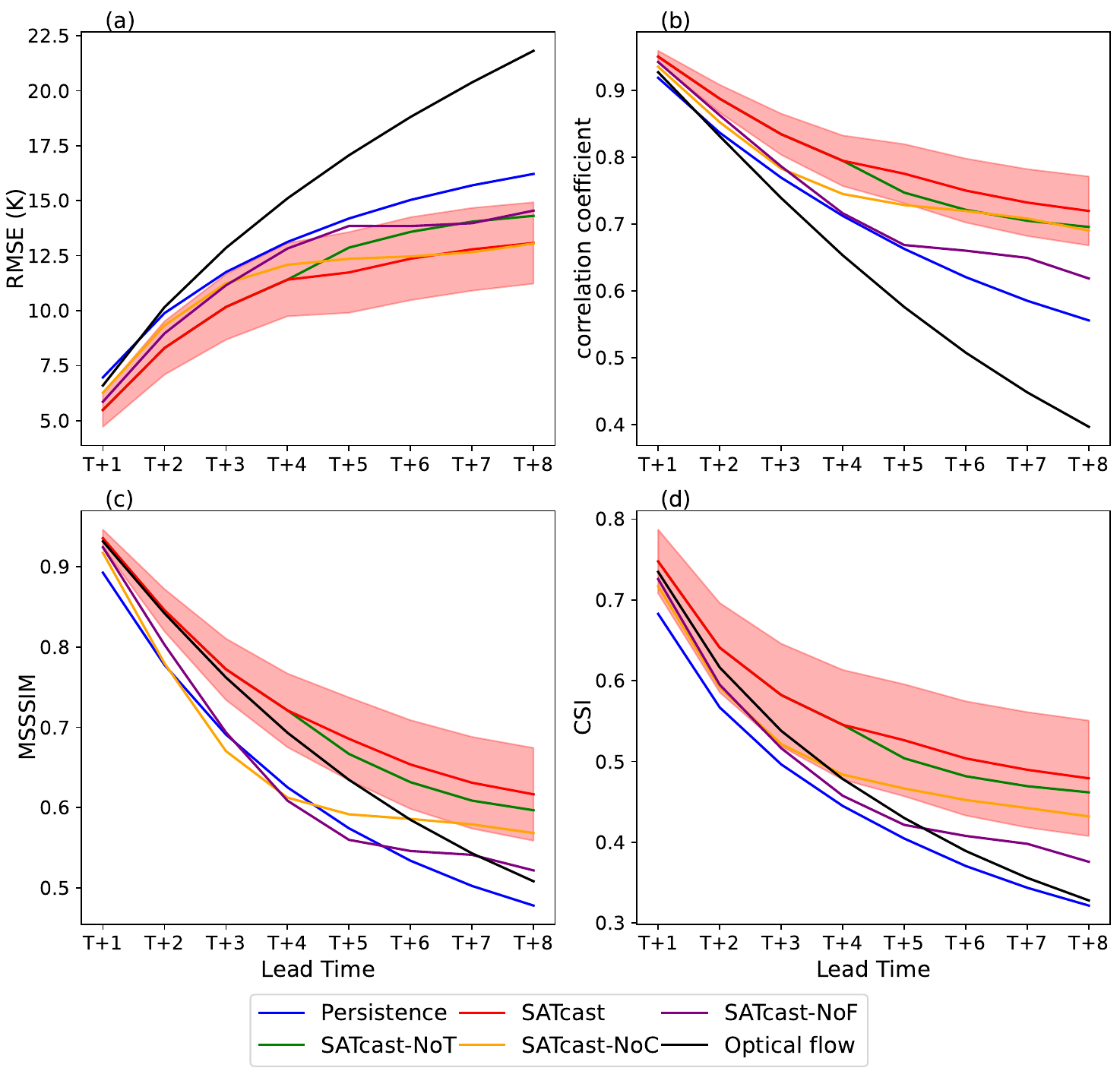}
\caption{Comparison of RMSE, correlation coefficient, MSSSIM, and CSI spatially averaged over (86$^\circ$ to 150$^\circ$ E in longitude and 1$^\circ$ to 41$^\circ$ N in latitude across five models: Persistence, SATcast, SATcast-NoC, SATcast-NoT, SATcast-NoF, and optical flow. The results are based on testing data from September to December in 2022. A threshold of 240 K is used in the calculation of ACC. The red shading represents the one standard deviation range of SATcast.} \label{fig2}
\end{figure}

\subsection{Ablation experiments and nowcasting skills}\label{sec21}

In this subsection, we compare three variants of SATcast (see details in Table \ref{t3}) for forecast lead times up to 8 hours, using 1,024 sequences, each spanning 16 hours (with the first 8 hours as input and the subsequent 8 hours as the target). The sequences, collected from September to December 2022, are used to evaluate the effectiveness of various modules within SATcast. 

Figure \ref{fig2} presents the metrics, spatially averaged over the region of interest (86$^\circ$ to 150$^\circ$ E in longitude and 1$^\circ$ to 41$^\circ$ N in latitude), for the persistence model, SATcast, and the three SATcast variants.
CSI is still computed using a threshold of 240 K. Among all the variants, SATcast-NoC, which directly predicts the entire 8-hour sequence, performs the worst for lead times T+1 to T+3.
This is likely due to the backward propagation of errors from frames beyond T+4, which hinders its optimization for earlier predictions.
However, SATcast-NoC outperforms SATcast-NoF, which excludes FuXi forecasts for lead times beyond T+4, suggesting that FuXi forecasts enhance performance, as satellite imagery alone are insufficient for predicting longer lead times.

SATcast-NoF performs well up to T+3 but deteriorates rapidly thereafter, likely due to its inability to capture the physical processes underlying convective cloud evolution.
To further examine the impact of fine-tuning on SATcast predictions beyond the initial four hours, a variant without fine-tuning, SATcast-NoT, performs better than other models except SATcast. Fine-tuning is essential in aligning SATcast's predictions with the data distribution of satellite imagery, as SATcast-NoT often overestimates values beyond T+4 (Supplementary Figure S8).

\begin{figure}[t]
\centering
\noindent\includegraphics[width=\linewidth]{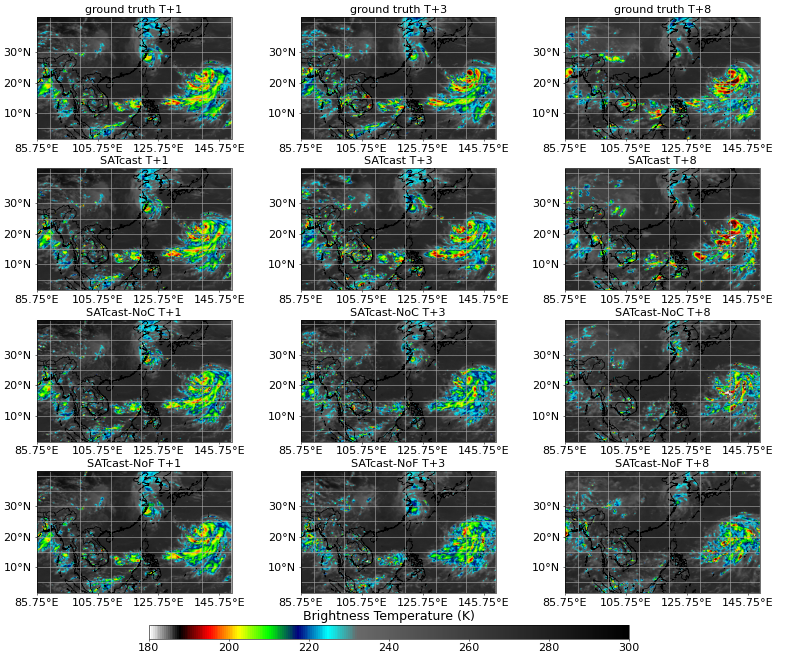}
\caption{Spatial distribution of brightness temperature at T+1, T+3, and T+8 from observation (first row), SATcast (second row), SATcast-NoC (third row), and SATcast-NoF predictions (fourth row), where T denotes the starting time of the prediction, the UTC time of T is 2022-09-14-06:00.} \label{fig3}
\end{figure}

Although the error differences among models appear subtle, Figure \ref{fig3} and Supplementary Figures S4-S7 reveal significant variations in their ability to predict the spatial distribution of convection, with SATcast showing the best skill. Figure \ref{fig3} also shows SATcast’s performance in predicting the intensification of Typhoon Nanmadol and the dissipation of Typhoon Muifa on September 13, 2022.
Muifa, one of the most powerful typhoons on record to strike Shanghai, and, Nanmadol, one of the strongest typhoons of 2022, caused substantial damage, undersocring the importance of accurate forecasts.

Starting at T+1 (2022-09-14-07:00 UTC), SATcast and its variants capture the textural features and central position of Nanmadol, characterized by a low cloud-top temperature and distinct core region.
However, significant differences emerge by T+3 (2022-09-14-09:00 UTC).
Without FuXi forecasts, SATcast-NoF demonstrates significant degradation in its ability to resolve the typhoon's structure, and SATcast-NoC produces only a loosely organized system.
In contrast, SATcast maintains robust skill in delineating both the eye of the typhoon and its structural integrity (further details are provided in Supplementary Figure S7).
By T+8 (2022-09-14-14:00 UTC), these differences become more pronounced.
SATcast-NoC and SATcast-NoF predict only fragmented and sporadic convection patterns, while SATcast successfully captures the intensification and movement of the tropical cyclone.
Although SATcast exhibits some discrepancies in the detailed organization of convection compared to observations, it remains the most reliable model in maintaining the typhoon's overall structure and dynamics.

As Typhoon Muifa approached the east coast of China, it gradually weakened and split into two cloud clusters by T+3.
SATcast demonstrated superior skill in accurately capturing both the spatial evolution and intensity variations during this process.
By T+8, SATcast's predictions exhibit a slight southward bias in the location of convective clouds north of Shanghai compared to observations. 
Despite this, SATcast still accurately forecasts the variations in convection intensity and shape.
In contrast, although SATcast-NoC and SATcast-NoF also predict a weakening system, their representations of convective cloud organization show significant morphological inaccuracies. 

Notably, satellite observations indicate that scattered convective clouds over Southeast Asia and the South China Sea intensified over time.
SATcast accurately predicts this strengthening at T+3, and its forecast at T+8 remains acceptable.
In comparison, both SATcast-NoC and SATcast-NoF significantly underestimate the intensity of these convective clouds at T+3 and T+8. 
Additional cases, including those with and without tropical cyclones, are detailed in Supplementary Figures S4-S7.
In these cases, SATcast consistently outperforms its variants in predicting changes in the intensity and organization of convective clouds, while the variants often erroneously predict weakening and dispersion of convective systems.

\subsection{Physical interpretations}\label{sec23}

Permutation feature importance is a valuable technique for interpreting multimodal deep learning models, especially in atmospheric studies \cite{joshi2021review, lagerquist2021using}.
This method quantifies the importance of each feature by selectively shuffling specific feature dimensions and measuring the resulting performance degradation.
It helps identify key physical parameters for predicting satellite imagery and refining the model.
To address the computational cost of repeated shuffling, we propose a threshold-based permutation strategy (detailed in Supplementary Information).
This strategy enables effective shuffling with only one shuffle per batch, significantly reducing computational costs.

Figure 5 presents heatmaps of the normalized mean squared error (N-MSE) ratios between permuted and baseline predictions ($\frac{MSE_{f}}{MSE_{m}}$).
Higher ratios correspond to greater feature importance, highlighting the most influential features for model predictions.

\begin{figure}[!t]
\centering
\noindent\includegraphics[width=\linewidth]{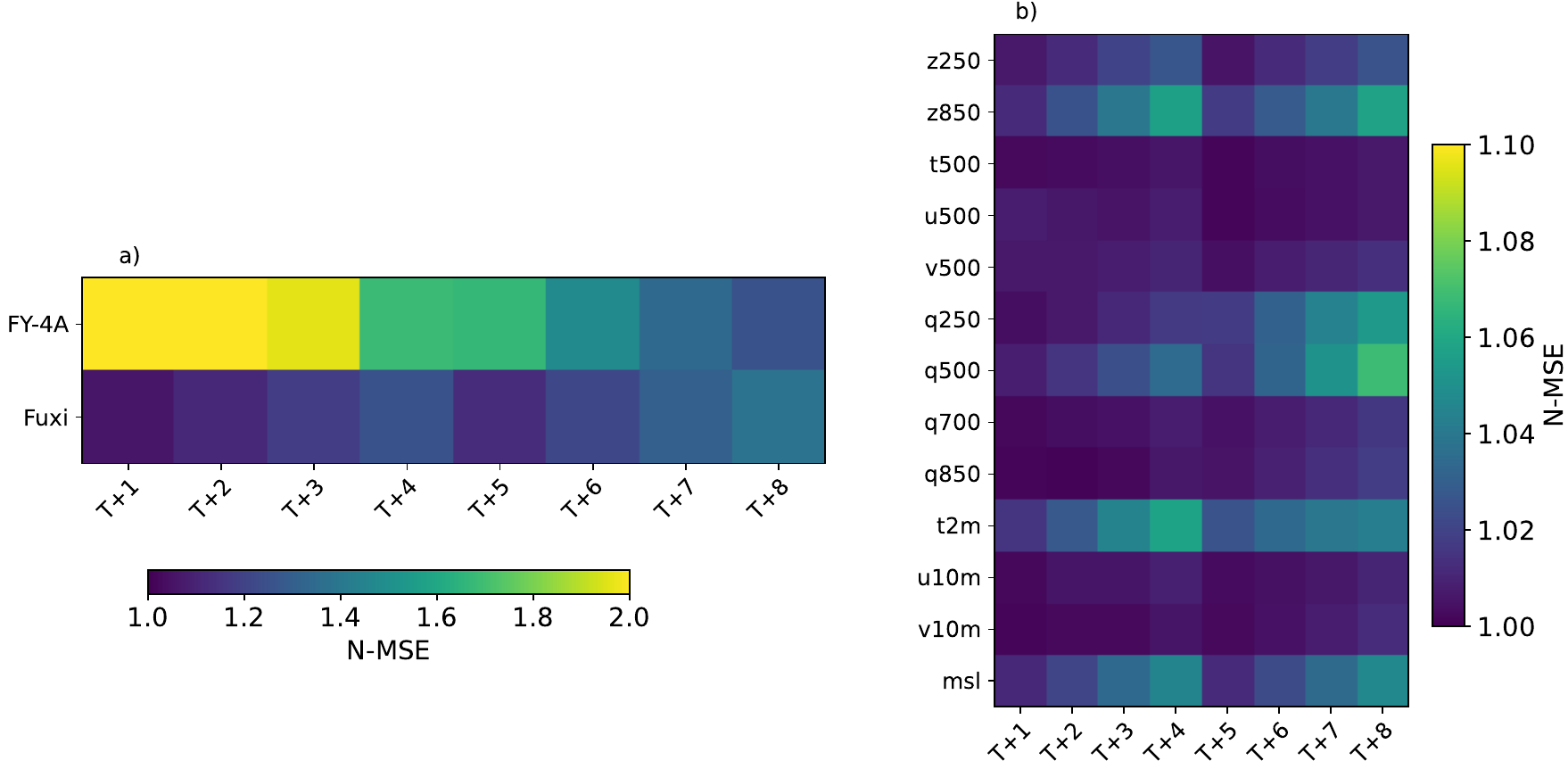}
\caption{Heatmap of N-MSE from the permutation feature importance. The x-axis represents the different features, and the y-axis represents the lead time. Panels are divided into a) higher importance and b) lower importance. FY-4A denotes the satellite imagery, and FuXi represents all 13 variables shown in the right panel. The three-digits codes on the y-axis of the right figure indicate pressure levels: 250 hPa, 500 hPa, 700 hPa, and 850 hPa. Variable abbreviations are as follows: z (geopotential), t (temperature), u (u component of wind), v (v component of wind), q (specific humidity), t2m (temperature at 2 meters), u10m (u component of wind at 10 meters), v10 (v component of wind at 10 meters), msl (mean sea-level pressure).} \label{fig6}
\end{figure}

\begin{figure}[h]
\centering
\noindent\includegraphics[width=0.7\linewidth]{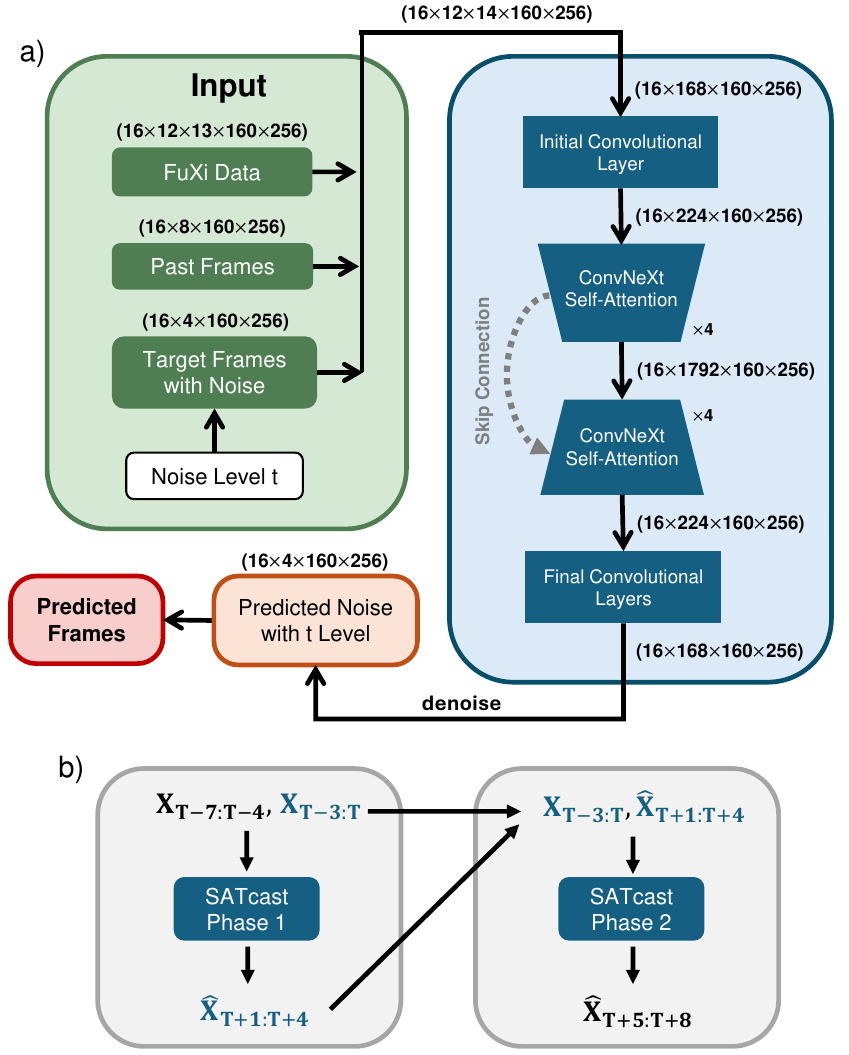}
\caption{a) SATcast architecture: dashed lines indicate skip connections. The denoising procedure uses the network highlighted in the blue box. The variations of dimensions are indicated when the sequence has 12 hours. b) Training and inference processes of SATcast: "T" represents the final time point of the input satellite image sequence. The SATcast-phase 1 predicts the satellite imagery from T+1 to T+4, the SATcast-phase 2 used predicted satellite imageries from SATcast-phase 1 and other past frames and FuXi data to predict satellite imagery from T+5 to T+8.}
\label{fignet}
\end{figure}

Figure \ref{fig6}a shows the feature importance, measured by N-MSE variations, for the eight hours of historical satellite imagery and FuXi predicted variables.
For T+1 to T+3, the N-MSE values from shuffled satellite imagery are significantly higher than those from the FuXi data, indicating the essential role of satellite observations during this period.
However, as the lead time increases, the importance of satellite imagery gradually diminishes, while the significance of FuXi data grows, ultimately surpassing that of satellite data at T+8.
This shift highlights the increasing relevance of FuXi data, which provides insights into future atmospheric patterns, for predictions over longer lead times.
It also explains why SATcast-NoC outperforms SATcast-NoF after T+3.

Figure \ref{fig6}b examines the importance of individual FuXi variables, all of which exhibit relatively low N-MSE values (below 1.1), with several parameters emerging as particularly impactful. Variables across different pressure levels, including high (250 hPa), middle (500 and 700 hPa), and low (850 hPa), are selected to ensure comprehensive representation.
Notably, the importance of these variables varies with forecast lead time.
Key features generating N-MSE about 1.05 include geopotential at 250 and 850 hPa (z250 and z850), specific humidity at 250 and 500 hPa (q250 and q500), near-surface temperature (t2m), and mean sea-level pressure (msl).
z250 and z850 influences atmospheric motions, such as convergence and divergence patterns, through geostrophic balance and adjustment, thereby guiding cloud movement and intensity changes.
q250 and q500, which are closely tied to clouds, exhibit a more pronounced influence from T+5 to T+8, whereas lower-level specific humidity (q850) has less impact.
Surface temperature and pressure contribute thermodynamic and dynamic forcing to the atmosphere, affecting prediction accuracy.
Furthermore, msl and geopotential (z250 and z850) serve as practical indicators of convection centers and tropical cyclones.
These findings demonstrate that incorporating critical atmospheric fields significantly enhances the performance of the SATcast models.

The results obtained by SATcast on channel 9 (Supplementary Figures S9-S10) further demonstrate the model's capacity to interpret physical information from a different perspective.
Brightness temperature on Channel 9, which represent high-level water vapor, exhibit greater stability, leading to superior performance across various metrics compared to those from channel 12.
This distinction can be attributed to the fact that, for shorter lead times, SATcast operates more like a video prediction model, heavily relying on satellite imagery for prediction.
However, for longer lead times, the model's performance declines rapidly due to inconsistencies between the targets detected by channels 9 (high-level water vapor) and 12 (cloud).
This mismatch hinders the model's ability to accurately interpret the influence of physical variables on the target, resulting in a significant reduction in skill for channel 9 predictions.

\section{Discussion }\label{sec5}

In this study, we introduce SATcast, a cascade, autoregressive, and multimodal deep learning model developed for nowcasting convective clouds.
SATcast combines atmospheric physical conditions predicted by FuXi with FY-4A satellite observations to generate high-fidelity cloud forecasts using a diffusion model.
SATcast effectively addresses long-standing challenges in nowcasting, such as image blurring and the rapid decline of forecast accuracy over time. 
The model produces high-quality cloud fields and accurately forecasts both the temporal and spatial characteristics of clouds up to 24 hours in advance.

SATcast adopts a cascade framework, with SATcast-phase 1 providing forecasts up to four hours ahead, while the SATcast-phase 2 takes SATcast-phase 1's outputs as input to extend predictions beyond the initial four hours.
SATcast-phase 1 and SATcast-phase 2 are optimized for lead times of 1 to 4 hours and 5 to 8 hours, respectively.
Remarkably, the model’s predictive capabilities extend beyond the optimized 8-hour forecast, as demonstrated in the 8-hour and 24-hour forecast tests, where it exhibited robustness and low cumulative error.
Even with a 24-hour lead time, the forecasts remain valid and reliable. 
Moreover, SATcast can be applied to a different satellite channel from the same satellite without significant performance degradation.
The model's robust performance probably benefits from its incorporation of atmospheric physical information, the cascade structure, and the stability and generative power of the diffusion model.

However, one limitation in this study is the loss of information due to downsampling satellite observations to a coarser resolution of 0.25$^{\circ}$ to reduce computational complexity.
In the future, we aim to develop latent-space diffusion models to mitigate these computational demands and enable nowcasting at FY satellite’s native 4-km resolution, thereby capturing more detailed storm features.
Another potential direction for future work is the generatation of probabilistic forecasts.
Convective systems exhibit relatively low predictability, and deterministic forecasts cannot quantify uncertainty, which is critical for making informed, contingent decisions.
The recent development of GenCast by Google Deepmind highlights the potential of diffusion models for producing ensemble forecasts by introducing additional stochasticity \cite{price_probabilistic_2024}.
Future convective cloud nowcasting could adopt a similar approach to provide more reliable probabilistic storm predictions based on satellite data.

To our knowledge, SATcast is the first model capable of accurately forecasting satellite imagery beyond 8 hours, and even up to 24 hours.
SATcast can play a critical role in providing timely alerts to vulnerable regions that lack advanced observation networks and high-performance computing facilities, as it can be operated without relying on NWP forecasts.
It is also vital for aviation, especially for flights over oceans and remote areas where ground-based radar observations are inaccessible, and it can effectively track typhoon paths over the ocean.
Additionally, cloud coverage is crucial for managing power grids and optimizing energy storage in solar photovoltaic plants \cite{xia2024accurate}.

\section*{Methods}\label{sec11}

\subsection*{Data}\label{sec110}

We obtained the sequences by incorporating 13 physical variables (detailed in Supplementary Table 1) from FuXi's hourly predictions, aligned with the spatiotemporal coordinates of the FY-4A satellite data.

Our analysis utilizes FY-4A AGRI Level 1 data from channel 12, which has a central wavelength of 10.7 µm, a spatial resolution of 4 km, and a temporal resolution of 1 hour, covering the period from 2019 to 2022. This channel primarily captures cloud top temperature and surface temperature.
The data is then interpolated to a $0.25^{\circ}$ resolution to reduce computational costs while maintaining resolution consistent with FuXi data.

These sequences are combined with the 13 physical variables predicted by FuXi every 12 hours, aligned with the latitude, longitude, and time information from the satellite data. The resulting dataset is organized in a T, C, H, W format (12, 14, 160, 256), where T represents time steps, C denotes channels (including both satellite data and physical variables), and H and W correspond to the spatial dimensions in the latitude and longitude directions. 
The training dataset for SATcast and SATcast-NoF comprises 19,904 sequences, each spanning 12 hours. Additionally, a separate dataset for training SATcast-NoC includes 17,408 sequences, each with a duration of 16 hours. To evaluate the performance of SATcast and assess the impact of its modules, distinct validation sequences lasting 16 and 32 hours were also prepared.

\subsection*{Basic diffusion}\label{sec111}

In the forward process of the diffusion model, noise is incrementally added to the target.
Let $P(x_{0})$ represent the sample distribution.
The following discrete-time Gaussian process is defined:

\begin{equation}\label{e1}
q(x_{t}|x_{t-1})=N(x_{t};\sqrt{1-\beta _{t}}x_{t-1},\beta _{t}\textit{\textbf{I}})
\end{equation}
where $x_{t}$ denotes the noise added at time step $t$ to $x_{0}$, and is calculated as: 
$x_{t} =\sqrt{\bar{\alpha_{t}}}x_{0}+\sqrt{1-\bar{\alpha _{t}}}\epsilon$, with $\epsilon$ being the noise sampled from $\mathcal{N}(0 \sim 1)$.
Here, $\alpha_{t}$ is a fixed schedule over $t$, and $\beta_{t}=1-\alpha_{t}$.
The network is trained to learn the noise injected to the data.
During sampling, the diffusion model reverses the forward diffusion process to recover $x_{t-1}$ from $x_{t}$, given by:

\begin{equation}\label{e2}
p(x_{t-1}|x_{t},x_{0})=\mathcal{N}(x_{t-1};\tilde{\mu _{t}}(x_{t},x_{0}),\tilde{\beta _{t}}\mathbf{I})
\end{equation}
where $\tilde{\mu _{t}}(x_{t},x_{0}=\frac{1}{\sqrt{\alpha _{t}}}(x_{t}-\frac{1-\alpha _{t}}{\sqrt{1-\bar{\alpha _{t}}}}\epsilon _{\theta })$, $\bar{\alpha _{t}}=\prod_{1}^{t}\alpha _{t}$, and $\epsilon _{\theta }$ is the noise predicted by the network.
Additionally, $\tilde{\beta _{t}}=\frac{1-\bar{\alpha }_{t-1}}{1-\bar{\alpha }_{t}}\beta _{t}$.
Thus, $x_{t-1}$ can be sampled from the mean and variance \cite{dhariwal2021diffusion,nichol2021improved}.

\subsection*{Conditional diffusion for satellite image nowcasting}\label{sec112}

Consider a sequence of satellite image time series, where the nowcasting task predicts future $y$ frames satellite imageries based on a past $x$ frames satellite imageries and FuXi variables from time steps $\mathrm{T}-x-1$ to $\mathrm{T}+y$. During training, noise is added to the future $y$ frames at level $t$, while the past frames and FuXi variables serve as the conditions.
Therefore, the reverse process is modified as:

\begin{equation}\label{e3}
p(x_{t-1}|x_{t},x_{0},cond)=\mathcal{N}(x_{t-1};\tilde{\mu _{t}}(x_{t},x_{0},cond),\tilde{\beta _{t}}\mathbf{I})
\end{equation}

The network training objective is simplified as follows:

\begin{equation}\label{e4}
\mathcal L=\mathbb{E}_{x_{t},cond,t,\epsilon \sim \mathcal N (0,1)} \left\|\epsilon -\epsilon _{\theta }(x_{t},cond,t) \right\|_{2}^{2}
\end{equation}
where $\epsilon _{\theta }$ represents the noise predicted by the network.

\subsection*{Model}\label{sec113}

The basic model, SATcast, is a cascade model that incorporates FuXi data, and undergoes a two-phase training, as illustrated in Figure 6b.
In the first phase, the model learns to predict frames from T+1 to T+4 using past satellite frames from T$-$7 to T+0 and the corresponding FuXi predictions from T$-$7 to T+4.
The second phase extends the prediction to frames T+5 to T+8, utilizing both past satellite frames from T$-$3 to T+0, the previously generated frames from T+1 to T+4, and the corresponding FuXi predictions.
To optimize computational efficiency, SATcast-phase 2 is initialized with weights from SATcast-phase 1. We also present results from SATcast-NoT, where SATcast did not undergo fine-tuning in phase 2.
The second variant, SATcast-NoC, adopts a direct prediction approach, predicting frames T+1 to T+8 simultaneously using the same input data as SATcast-phase 1.
The third variant, SATcast-NoF excludes FuXi predictions as conditions while maintaining autoregressive prediction for all eight frames.

\begin{table}[h]
    \caption{The configurations of the SATcast and variants. NoC: No cascade (no autoregressive and fine-tuning), NoF: No FuXi data, NoT: No fine-tuning}\label{t3}
    \centering

    \begin{tabular}{llll}
    \hline\hline
    Model&Autoregressive&Fine-Tuning&FuXi data\\
\hline
SATcast& \Checkmark &\Checkmark&\Checkmark\\
SATcast-NoC&\ding{55}&\ding{55}&\Checkmark\\
SATcast-NoT&\Checkmark&\ding{55}&\Checkmark\\
SATcast-NoF&\Checkmark&\ding{55}&\ding{55}\\
    \hline
    \end{tabular}
    \end{table}

\subsubsection*{Cascade structure}\label{sec113}

We designed a cascade structure to allow the model to generate future convective clouds in multiple steps. Since predicting frames with longer lead times is more challenging, enabling the model to focus on earlier frames in a single forecast is more beneficial for training an effective model. Fine-tuning can be applied in subsequent iterations ti effectively reduce iterative errors \cite{chen2023fuxi}.
Additionally, Supplementary Figure S11 shows a sample from the training of SATcast-NoC, where SATcast-NoC begins to capture the structure of the typhoon in the first four frames but fails to predict the subsequent four frames.
This illustrates the rationale behind designing the cascade model to forecast four frames at a time.


\subsubsection*{Multimodal data}\label{sec113}

The physical-driven network is constructed by incorporating the physical variables from FuXi forecasts. 
We treat data representing the same variable at different altitudes as distinct modalities.
Consequently, we apply normalization to these disparate modalities by linearly interpolating their values within the range between -1 and 1.
After temporal and spatial alignment, we obtain a five-dimensional dataset characterized by dimensions B, T, C, H, W.
For a sequence of 12 time steps, the dimensions are 16, 12, 14, 160, 256.
Considering the fact that all variables can be expressed within a system of atmospheric dynamic equations, and to mitigate the loss from encoding compression and computational resources \cite{voleti2022mcvd}, the T and C dimensions within the five-dimensional data are merged before entering the model, resulting in a structure of B, T$\times$C, H, W.
This is followed by a convolutional layer with a larger kernel size to extract spatial information. 
We believe this approach enhances the model's ability to accurately obtain the conditional information, ensuring that the predicted satellite imagery maintain correct spatial characteristics rather than merely presenting enhanced quality.

\subsubsection*{Network}\label{sec113}
We employed a U-Net2D architecture to predict noise, as shown in Figure 6a.
This U-Net contains ConvNeXt modules \cite{liu2022convnet} and self-attention modules \cite{vaswani2017attention}.
Notably, the network does not include specific modules for processing the temporal correlations of the data.
Instead, the temporal characteristics are implicitly learned by the model.

As previously described, after the initial convolutional layer, the data were reshaped to dimensions 16, 224, 160, 256.
The data then pass through four downsampling blocks, where both the H and W are halved, while the number of channels is doubled.
Each downsampling block consists of two ConvNeXt and self-attention modules with residual connections.
Following this, the data undergo four upsampling blocks of the same structure with the downsampling blocks, to restore the original dimensions before entering the UNet.
Finally, multiple convolutional layers progressively reduce the number of channels to T$\times$C (12$\times$14).

At the end of the network, the dimensions are restored, and only the noise-added portion, the target satellite imagery, is extracted to calculate the loss with the added noise.

\subsection*{Model training}\label{sec114}

The model is developed using Pytorch \cite{Paszke2017}.
Training the model takes approximately 20 hours on a cluster of 8 Nvidia H800 GPUs.
The model is trained for 300 epochs using a batch size of 16 per GPU.
The AdamW \cite{loshchilov2017decoupled} optimizer is used with parameters $\beta_{1}=0.9$ and $\beta_{2}=0.99$, and the learning rate is initialized at $5.0\times 10^{-5}$ with a warm-up phase followed by cosine annealing.
An exponential moving average is also applied during training \cite{morales2024exponential}.
The smooth L1 loss is used as the loss function. 

To improve the model's ability to predict severe cases, sequences containing category three typhoons are resampled once.
Classifier-free guidance is employed during both training and sampling to improve image quality \cite{ho2022classifier}. Additionally, offset noise is applied during sampling to reduce mismatches in data distribution \cite{lin2024commondiffusionnoiseschedules}.

\subsection*{Evaluation method}\label{sec115}
For target regions spanning a wide range of latitude, we apply latitude-weighted RMSE, correlation coefficient and CSI, which are calculated as follows:

\begin{equation}\label{e6}
\mathcal  
\mathbf{\mathbf{RMSE}}(\tau )=\sqrt{\frac{1}{H\times W}\sum_{i=1}^{H}\sum_{j=1}^{W}a_{i}(\mathbf{\hat{X}}_{i,j}^{\mathrm{T_{0}}+\tau}-\mathbf{X}_{i,j}^{\mathrm{T_{0}}+\tau})^{2}}
\end{equation}

\begin{equation}\label{e7}
\mathcal  
\rho(\tau)=\frac{1}{H\times W}\sum_{i=1}^{H}\sum_{j=1}^{W}a_{i}\frac{\mathbf{cov}(\mathbf{\hat{X}}^{\mathrm{T_{0}}+\tau}_{i,j},\mathbf{X}^{\mathrm{T_{0}}+\tau}_{i,j})}{\sigma _{\mathbf{\hat{X}}^{\mathrm{T_{0}}+\tau}_{i,j}}{\sigma _{\mathbf{X}^{\mathrm{T_{0}}+\tau}_{i,j}}}}
\end{equation}

\begin{equation}\label{e8}
\mathcal  
\mathbf{\mathbf{CSI}}(\tau)=\frac{1}{H\times W}\sum_{i=1}^{H}\sum_{j=1}^{W}a_{i}\frac{TP^{\mathrm{T_{0}}+\tau}_{i,j}}{TP^{\mathrm{T_{0}}+\tau}_{i,j}+TN^{\mathrm{T_{0}}+\tau}_{i,j}+FP^{\mathrm{T_{0}}+\tau}_{i,j}}
\end{equation}
where $\tau$ is the forecast lead time, $\mathrm{T_{0}}$ is the initial time, and H and W correspond to the numbers of grid points in longitude and latitude, respectively. $a_{i}$ is the latitude weighting factor for the ith latitude index \cite{Rasp2020WeatherBenchAB}:

\begin{equation}\label{e10}
\mathcal  
\mathbf{a}_{i}
=\frac{\textrm{cos}(\textrm{lat}(i))}
{\frac{1}{H}\sum_{i}^{H}\textrm{cos}(\textrm{lat}(i))}
\end{equation}

CSI is commonly used for the evaluation of severe weather forecasts. TP refers to the number of correctly predicted positive pixels; FP is the number of negative pixels incorrectly predicted as positive; FN is the number of positive pixels incorrectly predicted as negative; and TN is the number of correctly predicted negative pixels.

MSSSIM combines Structural Similarity Index (SSIM) at multiple scales to yield more robust results. Latitude-weighting is not applied here, as the image is treated as a whole. 
It is calculated by:

\begin{equation}\label{e11}
\mathcal  
\mathbf{\mathbf{MSSSIM}}(\tau)=\left [ l_{M}(\mathbf{\hat{X}}^{\mathrm{T_{0}}+\tau},\mathbf{X}^{\mathrm{T_{0}}+\tau}) \right ]^{\alpha _{M}}\prod_{j=1}^{M}\left [ c_{j}(\mathbf{\hat{X}}^{\mathrm{T_{0}}+\tau},\mathbf{X}^{\mathrm{T_{0}}+\tau}) \right ]^{\beta _{j}}\left [ s_{j}(\mathbf{\hat{X}}^{\mathrm{T_{0}}+\tau},\mathbf{X}^{\mathrm{T_{0}}+\tau}) \right ]^{\gamma  _{j}}
\end{equation}

The comparisons about luminance, contrast and structure are denoted by the three terms in equation 9, the $M$ means difference scales of the images. The three exponents adjust the relative importance of different components. More details about the calculation can be found in Wang el al. (2020) \cite{1292216}.

\section*{Data availability}

The Fengyun-4A satellite imagery can be downloaded on \url{https://satellite.nsmc.org.cn/portalsite/default.aspx?currentculture=en-US}.

\section*{Code availability}
The scripts about the network, training, and inference can be found on \url{https://github.com/cd4tpcell/SATcast/tree/main} \cite{chen_2025_14643154}.

\backmatter

\section*{Acknowledgements}

The work described in this paper was substantially supported by a grant from the Research Grants Council (RGC) of the Hong Kong Special Administrative Region, China (Project Reference Number: AoE/P-601/23-N) and the Center for Ocean Research in Hong Kong and Macau (CORE), a joint research center between the Laoshan Laboratory and the Hong Kong University of Science and Technology (HKUST). HC, YH, and XS are additionally supported by RGC grant HKUST-16301322 and the Aviation Research and Development Project Phase 2 (AvRDP-2) of the World Meteorological Organization (WMO).
The authors thank the computing platform supported by HKUST Superpod Cluster to train the model and  the Artificial Intelligence Innovation and Incubation Institute of Fudan University for providing achives of the FuXi data.

\section*{Author contributions}
X.S., H.L., and X.L., conceptualized the project. H.C., and X.Z., completed the model training and evaluation, as well as the design of the experiments. Y.C., and P.C., provided the satellite data and helped with relevant analysis. Q.Z., contributed to the data cleansing and tested the workflow for model training. Y.H., helped with data analysis and visualization. H.C., X.Z., and X.S., prepared the initial manuscript. All authors contributed to further improvements.

\section*{Competing interests}

The authors declare no competing interest.

\bibliography{references}


\begin{thebibliography}{45}
\ifx \bisbn   \undefined \def \bisbn  #1{ISBN #1}\fi
\ifx \binits  \undefined \def \binits#1{#1}\fi
\ifx \bauthor  \undefined \def \bauthor#1{#1}\fi
\ifx \batitle  \undefined \def \batitle#1{#1}\fi
\ifx \bjtitle  \undefined \def \bjtitle#1{#1}\fi
\ifx \bvolume  \undefined \def \bvolume#1{\textbf{#1}}\fi
\ifx \byear  \undefined \def \byear#1{#1}\fi
\ifx \bissue  \undefined \def \bissue#1{#1}\fi
\ifx \bfpage  \undefined \def \bfpage#1{#1}\fi
\ifx \blpage  \undefined \def \blpage #1{#1}\fi
\ifx \burl  \undefined \def \burl#1{\textsf{#1}}\fi
\ifx \doiurl  \undefined \def \doiurl#1{\url{https://doi.org/#1}}\fi
\ifx \betal  \undefined \def \betal{\textit{et al.}}\fi
\ifx \binstitute  \undefined \def \binstitute#1{#1}\fi
\ifx \binstitutionaled  \undefined \def \binstitutionaled#1{#1}\fi
\ifx \bctitle  \undefined \def \bctitle#1{#1}\fi
\ifx \beditor  \undefined \def \beditor#1{#1}\fi
\ifx \bpublisher  \undefined \def \bpublisher#1{#1}\fi
\ifx \bbtitle  \undefined \def \bbtitle#1{#1}\fi
\ifx \bedition  \undefined \def \bedition#1{#1}\fi
\ifx \bseriesno  \undefined \def \bseriesno#1{#1}\fi
\ifx \blocation  \undefined \def \blocation#1{#1}\fi
\ifx \bsertitle  \undefined \def \bsertitle#1{#1}\fi
\ifx \bsnm \undefined \def \bsnm#1{#1}\fi
\ifx \bsuffix \undefined \def \bsuffix#1{#1}\fi
\ifx \bparticle \undefined \def \bparticle#1{#1}\fi
\ifx \barticle \undefined \def \barticle#1{#1}\fi
\bibcommenthead
\ifx \bconfdate \undefined \def \bconfdate #1{#1}\fi
\ifx \botherref \undefined \def \botherref #1{#1}\fi
\ifx \url \undefined \def \url#1{\textsf{#1}}\fi
\ifx \bchapter \undefined \def \bchapter#1{#1}\fi
\ifx \bbook \undefined \def \bbook#1{#1}\fi
\ifx \bcomment \undefined \def \bcomment#1{#1}\fi
\ifx \oauthor \undefined \def \oauthor#1{#1}\fi
\ifx \citeauthoryear \undefined \def \citeauthoryear#1{#1}\fi
\ifx \endbibitem  \undefined \def \endbibitem {}\fi
\ifx \bconflocation  \undefined \def \bconflocation#1{#1}\fi
\ifx \arxivurl  \undefined \def \arxivurl#1{\textsf{#1}}\fi
\csname PreBibitemsHook\endcsname

\bibitem[\protect\citeauthoryear{Chen et~al.}{2024}]{chm2024cit}
\begin{barticle}
\bauthor{\bsnm{Chen}, \binits{H.}},
\bauthor{\bsnm{Shi}, \binits{X.}},
\bauthor{\bsnm{Nie}, \binits{X.}},
\bauthor{\bsnm{Wang}, \binits{Y.}},
\bauthor{\bsnm{Leung}, \binits{C.Y.Y.}},
\bauthor{\bsnm{Cheung}, \binits{P.}},
\bauthor{\bsnm{Chan}, \binits{P.W.}}:
\batitle{Tropical aviation turbulence induced by the interaction between a jet
  stream and deep convection}.
\bjtitle{Journal of Geophysical Research: Atmospheres}
\bvolume{129}(\bissue{18}),
\bfpage{2024}--\blpage{040763}
(\byear{2024})
\doiurl{10.1029/2024JD040763}
{\href{https://arxiv.org/abs/https://agupubs.onlinelibrary.wiley.com/doi/pdf/10.1029/2024JD040763}{{https://agupubs.onlinelibrary.wiley.com/doi/pdf/10.1029/2024JD040763}}}.
\bcomment{e2024JD040763 2024JD040763}
\end{barticle}
\endbibitem

\bibitem[\protect\citeauthoryear{Guo et~al.}{2022}]{guo2022thunderstorm}
\begin{barticle}
\bauthor{\bsnm{Guo}, \binits{Y.}},
\bauthor{\bsnm{Zhong}, \binits{M.}},
\bauthor{\bsnm{Chen}, \binits{X.}},
\bauthor{\bsnm{Zhou}, \binits{Z.}},
\bauthor{\bsnm{Xu}, \binits{G.}},
\bauthor{\bsnm{Xu}, \binits{G.}},
\bauthor{\bsnm{Dong}, \binits{L.}}:
\batitle{A thunderstorm gale forecast method based on the objective
  classification and continuous probability}.
\bjtitle{Atmosphere}
\bvolume{13}(\bissue{8}),
\bfpage{1308}
(\byear{2022})
\end{barticle}
\endbibitem

\bibitem[\protect\citeauthoryear{Organization}{2023}]{wmoGuidelinesSatellitebased}
\begin{botherref}
\oauthor{\bsnm{Organization}, \binits{W.M.}}:
{G}uidelines for {S}atellite-based {N}owcasting in {A}frica ---
  library.wmo.int.
\url{https://library.wmo.int/records/item/58348-guidelines-for-satellite-based-nowcasting-in-africa}.
[Accessed 14-12-2024]
(2023)
\end{botherref}
\endbibitem

\bibitem[\protect\citeauthoryear{Schmit et~al.}{2017}]{schmit2017closer}
\begin{barticle}
\bauthor{\bsnm{Schmit}, \binits{T.J.}},
\bauthor{\bsnm{Griffith}, \binits{P.}},
\bauthor{\bsnm{Gunshor}, \binits{M.M.}},
\bauthor{\bsnm{Daniels}, \binits{J.M.}},
\bauthor{\bsnm{Goodman}, \binits{S.J.}},
\bauthor{\bsnm{Lebair}, \binits{W.J.}}:
\batitle{A closer look at the abi on the goes-r series}.
\bjtitle{Bulletin of the American Meteorological Society}
\bvolume{98}(\bissue{4}),
\bfpage{681}--\blpage{698}
(\byear{2017})
\end{barticle}
\endbibitem

\bibitem[\protect\citeauthoryear{Baker and Matthews}{2004}]{baker2004lucas}
\begin{barticle}
\bauthor{\bsnm{Baker}, \binits{S.}},
\bauthor{\bsnm{Matthews}, \binits{I.}}:
\batitle{Lucas-kanade 20 years on: A unifying framework}.
\bjtitle{International journal of computer vision}
\bvolume{56},
\bfpage{221}--\blpage{255}
(\byear{2004})
\end{barticle}
\endbibitem

\bibitem[\protect\citeauthoryear{Smith et~al.}{2024}]{smith2024evaluating}
\begin{barticle}
\bauthor{\bsnm{Smith}, \binits{J.}},
\bauthor{\bsnm{Birch}, \binits{C.}},
\bauthor{\bsnm{Marsham}, \binits{J.}},
\bauthor{\bsnm{Peatman}, \binits{S.}},
\bauthor{\bsnm{Bollasina}, \binits{M.}},
\bauthor{\bsnm{Pankiewicz}, \binits{G.}}:
\batitle{Evaluating pysteps optical flow algorithms for convection nowcasting
  over the maritime continent using satellite data}.
\bjtitle{Natural Hazards and Earth System Sciences}
\bvolume{24}(\bissue{2}),
\bfpage{567}--\blpage{582}
(\byear{2024})
\end{barticle}
\endbibitem

\bibitem[\protect\citeauthoryear{Zhang et~al.}{2023}]{zhang2023skilful}
\begin{barticle}
\bauthor{\bsnm{Zhang}, \binits{Y.}},
\bauthor{\bsnm{Long}, \binits{M.}},
\bauthor{\bsnm{Chen}, \binits{K.}},
\bauthor{\bsnm{Xing}, \binits{L.}},
\bauthor{\bsnm{Jin}, \binits{R.}},
\bauthor{\bsnm{Jordan}, \binits{M.I.}},
\bauthor{\bsnm{Wang}, \binits{J.}}:
\batitle{Skilful nowcasting of extreme precipitation with nowcastnet}.
\bjtitle{Nature}
\bvolume{619}(\bissue{7970}),
\bfpage{526}--\blpage{532}
(\byear{2023})
\end{barticle}
\endbibitem

\bibitem[\protect\citeauthoryear{Chen et~al.}{2023}]{chen2023fuxi}
\begin{barticle}
\bauthor{\bsnm{Chen}, \binits{L.}},
\bauthor{\bsnm{Zhong}, \binits{X.}},
\bauthor{\bsnm{Zhang}, \binits{F.}},
\bauthor{\bsnm{Cheng}, \binits{Y.}},
\bauthor{\bsnm{Xu}, \binits{Y.}},
\bauthor{\bsnm{Qi}, \binits{Y.}},
\bauthor{\bsnm{Li}, \binits{H.}}:
\batitle{Fuxi: A cascade machine learning forecasting system for 15-day global
  weather forecast}.
\bjtitle{npj Climate and Atmospheric Science}
\bvolume{6}(\bissue{1}),
\bfpage{190}
(\byear{2023})
\end{barticle}
\endbibitem

\bibitem[\protect\citeauthoryear{Bi et~al.}{2023}]{bi2023accurate}
\begin{barticle}
\bauthor{\bsnm{Bi}, \binits{K.}},
\bauthor{\bsnm{Xie}, \binits{L.}},
\bauthor{\bsnm{Zhang}, \binits{H.}},
\bauthor{\bsnm{Chen}, \binits{X.}},
\bauthor{\bsnm{Gu}, \binits{X.}},
\bauthor{\bsnm{Tian}, \binits{Q.}}:
\batitle{Accurate medium-range global weather forecasting with 3d neural
  networks}.
\bjtitle{Nature}
\bvolume{619}(\bissue{7970}),
\bfpage{533}--\blpage{538}
(\byear{2023})
\end{barticle}
\endbibitem

\bibitem[\protect\citeauthoryear{Kumar et~al.}{2020}]{kumar2020convcast}
\begin{barticle}
\bauthor{\bsnm{Kumar}, \binits{A.}},
\bauthor{\bsnm{Islam}, \binits{T.}},
\bauthor{\bsnm{Sekimoto}, \binits{Y.}},
\bauthor{\bsnm{Mattmann}, \binits{C.}},
\bauthor{\bsnm{Wilson}, \binits{B.}}:
\batitle{Convcast: An embedded convolutional lstm based architecture for
  precipitation nowcasting using satellite data}.
\bjtitle{Plos one}
\bvolume{15}(\bissue{3}),
\bfpage{0230114}
(\byear{2020})
\end{barticle}
\endbibitem

\bibitem[\protect\citeauthoryear{Tran and Song}{2019}]{tran2019computer}
\begin{barticle}
\bauthor{\bsnm{Tran}, \binits{Q.-K.}},
\bauthor{\bsnm{Song}, \binits{S.-k.}}:
\batitle{Computer vision in precipitation nowcasting: Applying image quality
  assessment metrics for training deep neural networks}.
\bjtitle{Atmosphere}
\bvolume{10}(\bissue{5}),
\bfpage{244}
(\byear{2019})
\end{barticle}
\endbibitem

\bibitem[\protect\citeauthoryear{Ehsani et~al.}{2021}]{ehsani2021nowcasting}
\begin{botherref}
\oauthor{\bsnm{Ehsani}, \binits{M.R.}},
\oauthor{\bsnm{Zarei}, \binits{A.}},
\oauthor{\bsnm{Gupta}, \binits{H.V.}},
\oauthor{\bsnm{Barnard}, \binits{K.}},
\oauthor{\bsnm{Behrangi}, \binits{A.}}:
Nowcasting-nets: Deep neural network structures for precipitation nowcasting
  using imerg.
arXiv preprint arXiv:2108.06868
(2021)
\end{botherref}
\endbibitem

\bibitem[\protect\citeauthoryear{Wei
  et~al.}{2024}]{wei2024dayudatadrivenmodelgeostationary}
\begin{botherref}
\oauthor{\bsnm{Wei}, \binits{X.}},
\oauthor{\bsnm{Zhang}, \binits{F.}},
\oauthor{\bsnm{Zhang}, \binits{R.}},
\oauthor{\bsnm{Li}, \binits{W.}},
\oauthor{\bsnm{Liu}, \binits{C.}},
\oauthor{\bsnm{Guo}, \binits{B.}},
\oauthor{\bsnm{Li}, \binits{J.}},
\oauthor{\bsnm{Fu}, \binits{H.}},
\oauthor{\bsnm{Tang}, \binits{X.}}:
DaYu: Data-Driven Model for Geostationary Satellite Observed Cloud Images
  Forecasting
(2024).
\url{https://arxiv.org/abs/2411.10144}
\end{botherref}
\endbibitem

\bibitem[\protect\citeauthoryear{Shi et~al.}{2015}]{shi2015convolutional}
\begin{botherref}
\oauthor{\bsnm{Shi}, \binits{X.}},
\oauthor{\bsnm{Chen}, \binits{Z.}},
\oauthor{\bsnm{Wang}, \binits{H.}},
\oauthor{\bsnm{Yeung}, \binits{D.-Y.}},
\oauthor{\bsnm{Wong}, \binits{W.-K.}},
\oauthor{\bsnm{Woo}, \binits{W.-c.}}:
Convolutional lstm network: A machine learning approach for precipitation
  nowcasting.
Advances in neural information processing systems
\textbf{28}
(2015)
\end{botherref}
\endbibitem

\bibitem[\protect\citeauthoryear{Vaswani}{2017}]{vaswani2017attention}
\begin{botherref}
\oauthor{\bsnm{Vaswani}, \binits{A.}}:
Attention is all you need.
Advances in Neural Information Processing Systems
(2017)
\end{botherref}
\endbibitem

\bibitem[\protect\citeauthoryear{Wang et~al.}{2023}]{wang2023skillful}
\begin{botherref}
\oauthor{\bsnm{Wang}, \binits{R.}},
\oauthor{\bsnm{Su}, \binits{L.}},
\oauthor{\bsnm{Wong}, \binits{W.K.}},
\oauthor{\bsnm{Lau}, \binits{A.K.}},
\oauthor{\bsnm{Fung}, \binits{J.C.}}:
Skillful radar-based heavy rainfall nowcasting using task-segmented generative
  adversarial network.
IEEE Transactions on Geoscience and Remote Sensing
(2023)
\end{botherref}
\endbibitem

\bibitem[\protect\citeauthoryear{Goodfellow
  et~al.}{2020}]{goodfellow2020generative}
\begin{barticle}
\bauthor{\bsnm{Goodfellow}, \binits{I.}},
\bauthor{\bsnm{Pouget-Abadie}, \binits{J.}},
\bauthor{\bsnm{Mirza}, \binits{M.}},
\bauthor{\bsnm{Xu}, \binits{B.}},
\bauthor{\bsnm{Warde-Farley}, \binits{D.}},
\bauthor{\bsnm{Ozair}, \binits{S.}},
\bauthor{\bsnm{Courville}, \binits{A.}},
\bauthor{\bsnm{Bengio}, \binits{Y.}}:
\batitle{Generative adversarial networks}.
\bjtitle{Communications of the ACM}
\bvolume{63}(\bissue{11}),
\bfpage{139}--\blpage{144}
(\byear{2020})
\end{barticle}
\endbibitem

\bibitem[\protect\citeauthoryear{Kim et~al.}{2024}]{KIM2024105529}
\begin{barticle}
\bauthor{\bsnm{Kim}, \binits{W.}},
\bauthor{\bsnm{Jeong}, \binits{C.-H.}},
\bauthor{\bsnm{Kim}, \binits{S.}}:
\batitle{Improvements in deep learning-based precipitation nowcasting using
  major atmospheric factors with radar rain rate}.
\bjtitle{Computers \& Geosciences}
\bvolume{184},
\bfpage{105529}
(\byear{2024})
\doiurl{10.1016/j.cageo.2024.105529}
\end{barticle}
\endbibitem

\bibitem[\protect\citeauthoryear{Trier et~al.}{2014}]{trier2014use}
\begin{barticle}
\bauthor{\bsnm{Trier}, \binits{S.B.}},
\bauthor{\bsnm{Davis}, \binits{C.A.}},
\bauthor{\bsnm{Ahijevych}, \binits{D.A.}},
\bauthor{\bsnm{Manning}, \binits{K.W.}}:
\batitle{Use of the parcel buoyancy minimum (b min) to diagnose simulated
  thermodynamic destabilization. part i: Methodology and case studies of mcs
  initiation environments}.
\bjtitle{Monthly Weather Review}
\bvolume{142}(\bissue{3}),
\bfpage{945}--\blpage{966}
(\byear{2014})
\end{barticle}
\endbibitem

\bibitem[\protect\citeauthoryear{Shi and Wang}{2022}]{2022sxm}
\begin{botherref}
\oauthor{\bsnm{Shi}, \binits{X.}},
\oauthor{\bsnm{Wang}, \binits{Y.}}:
Impacts of cumulus convection and turbulence parameterizations on the
  convection-permitting simulation of typhoon precipitation.
Monthly Weather Review
\textbf{150}
(2022)
\doiurl{10.1175/MWR-D-22-0057.1}
\end{botherref}
\endbibitem

\bibitem[\protect\citeauthoryear{Radford et~al.}{2021}]{2021Learningclip}
\begin{botherref}
\oauthor{\bsnm{Radford}, \binits{A.}},
\oauthor{\bsnm{Kim}, \binits{J.W.}},
\oauthor{\bsnm{Hallacy}, \binits{C.}},
\oauthor{\bsnm{Ramesh}, \binits{A.}},
\oauthor{\bsnm{Goh}, \binits{G.}},
\oauthor{\bsnm{Agarwal}, \binits{S.}},
\oauthor{\bsnm{Sastry}, \binits{G.}},
\oauthor{\bsnm{Askell}, \binits{A.}},
\oauthor{\bsnm{Mishkin}, \binits{P.}},
\oauthor{\bsnm{Clark}, \binits{J.}}:
Learning transferable visual models from natural language supervision
(2021)
\end{botherref}
\endbibitem

\bibitem[\protect\citeauthoryear{Andrychowicz
  et~al.}{2023}]{Andrychowicz2023DeepLF}
\begin{botherref}
\oauthor{\bsnm{Andrychowicz}, \binits{M.}},
\oauthor{\bsnm{Espeholt}, \binits{L.}},
\oauthor{\bsnm{Li}, \binits{D.}},
\oauthor{\bsnm{Merchant}, \binits{S.}},
\oauthor{\bsnm{Merose}, \binits{A.}},
\oauthor{\bsnm{Zyda}, \binits{F.}},
\oauthor{\bsnm{Agrawal}, \binits{S.}},
\oauthor{\bsnm{Kalchbrenner}, \binits{N.}}:
Deep learning for day forecasts from sparse observations.
ArXiv
\textbf{abs/2306.06079}
(2023)
\end{botherref}
\endbibitem

\bibitem[\protect\citeauthoryear{Huang et~al.}{2016}]{2016Deep}
\begin{bchapter}
\bauthor{\bsnm{Huang}, \binits{G.}},
\bauthor{\bsnm{Sun}, \binits{Y.}},
\bauthor{\bsnm{Liu}, \binits{Z.}},
\bauthor{\bsnm{Sedra}, \binits{D.}},
\bauthor{\bsnm{Weinberger}, \binits{K.}}:
\bctitle{Deep networks with stochastic depth}.
In: \bbtitle{Springer International Publishing}
(\byear{2016})
\end{bchapter}
\endbibitem

\bibitem[\protect\citeauthoryear{Rui~Wang}{2024}]{2024wangdiff}
\begin{botherref}
\oauthor{\bsnm{Rui~Wang}, \binits{A.K.H.L.} \bsuffix{Jimmy C. H.~Fung}}:
Skillful precipitation nowcasting using physical-driven diffusion networks.
Geophysical Reasearch Letter
\textbf{51}(24)
(2024)
\end{botherref}
\endbibitem

\bibitem[\protect\citeauthoryear{Ho et~al.}{2020}]{ho2020denoising}
\begin{barticle}
\bauthor{\bsnm{Ho}, \binits{J.}},
\bauthor{\bsnm{Jain}, \binits{A.}},
\bauthor{\bsnm{Abbeel}, \binits{P.}}:
\batitle{Denoising diffusion probabilistic models}.
\bjtitle{Advances in neural information processing systems}
\bvolume{33},
\bfpage{6840}--\blpage{6851}
(\byear{2020})
\end{barticle}
\endbibitem

\bibitem[\protect\citeauthoryear{Dhariwal and
  Nichol}{2021}]{dhariwal2021diffusion}
\begin{barticle}
\bauthor{\bsnm{Dhariwal}, \binits{P.}},
\bauthor{\bsnm{Nichol}, \binits{A.}}:
\batitle{Diffusion models beat gans on image synthesis}.
\bjtitle{Advances in neural information processing systems}
\bvolume{34},
\bfpage{8780}--\blpage{8794}
(\byear{2021})
\end{barticle}
\endbibitem

\bibitem[\protect\citeauthoryear{Yang et~al.}{2017}]{yang2017introducing}
\begin{barticle}
\bauthor{\bsnm{Yang}, \binits{J.}},
\bauthor{\bsnm{Zhang}, \binits{Z.}},
\bauthor{\bsnm{Wei}, \binits{C.}},
\bauthor{\bsnm{Lu}, \binits{F.}},
\bauthor{\bsnm{Guo}, \binits{Q.}}:
\batitle{Introducing the new generation of chinese geostationary weather
  satellites, fengyun-4}.
\bjtitle{Bulletin of the American Meteorological Society}
\bvolume{98}(\bissue{8}),
\bfpage{1637}--\blpage{1658}
(\byear{2017})
\end{barticle}
\endbibitem

\bibitem[\protect\citeauthoryear{Lu et~al.}{2017}]{lu2017fy}
\begin{barticle}
\bauthor{\bsnm{Lu}, \binits{F.}},
\bauthor{\bsnm{Zhang}, \binits{X.-H.}},
\bauthor{\bsnm{Chen}, \binits{B.-Y.}},
\bauthor{\bsnm{Liu}, \binits{H.}},
\bauthor{\bsnm{Wu}, \binits{R.}},
\bauthor{\bsnm{Han}, \binits{Q.}},
\bauthor{\bsnm{Feng}, \binits{X.}},
\bauthor{\bsnm{Li}, \binits{Y.}},
\bauthor{\bsnm{Zhang}, \binits{Z.}}:
\batitle{Fy-4 geostationary meteorological satellite imaging characteristics
  and its application prospects}.
\bjtitle{J. Mar. Meteorol}
\bvolume{37}(\bissue{2}),
\bfpage{1}--\blpage{12}
(\byear{2017})
\end{barticle}
\endbibitem

\bibitem[\protect\citeauthoryear{Yan et~al.}{2024}]{yan2024study}
\begin{barticle}
\bauthor{\bsnm{Yan}, \binits{C.}},
\bauthor{\bsnm{Guang}, \binits{J.}},
\bauthor{\bsnm{Li}, \binits{Z.}},
\bauthor{\bsnm{Leeuw}, \binits{G.}},
\bauthor{\bsnm{Chen}, \binits{Z.}}:
\batitle{A study on typhoon center localization based on an improved
  spatio-temporally consistent scale-invariant feature transform and brightness
  temperature perturbations}.
\bjtitle{Remote Sensing}
\bvolume{16}(\bissue{21}),
\bfpage{4070}
(\byear{2024})
\end{barticle}
\endbibitem

\bibitem[\protect\citeauthoryear{Liu et~al.}{2022}]{liu2022convnet}
\begin{bchapter}
\bauthor{\bsnm{Liu}, \binits{Z.}},
\bauthor{\bsnm{Mao}, \binits{H.}},
\bauthor{\bsnm{Wu}, \binits{C.-Y.}},
\bauthor{\bsnm{Feichtenhofer}, \binits{C.}},
\bauthor{\bsnm{Darrell}, \binits{T.}},
\bauthor{\bsnm{Xie}, \binits{S.}}:
\bctitle{A convnet for the 2020s}.
In: \bbtitle{Proceedings of the IEEE/CVF Conference on Computer Vision and
  Pattern Recognition},
pp. \bfpage{11976}--\blpage{11986}
(\byear{2022})
\end{bchapter}
\endbibitem

\bibitem[\protect\citeauthoryear{Lagerquist et~al.}{2021}]{lagerquist2021using}
\begin{barticle}
\bauthor{\bsnm{Lagerquist}, \binits{R.}},
\bauthor{\bsnm{Stewart}, \binits{J.Q.}},
\bauthor{\bsnm{Ebert-Uphoff}, \binits{I.}},
\bauthor{\bsnm{Kumler}, \binits{C.}}:
\batitle{Using deep learning to nowcast the spatial coverage of convection from
  himawari-8 satellite data}.
\bjtitle{Monthly Weather Review}
\bvolume{149}(\bissue{12}),
\bfpage{3897}--\blpage{3921}
(\byear{2021})
\end{barticle}
\endbibitem

\bibitem[\protect\citeauthoryear{Wang et~al.}{2003}]{1292216}
\begin{bchapter}
\bauthor{\bsnm{Wang}, \binits{Z.}},
\bauthor{\bsnm{Simoncelli}, \binits{E.P.}},
\bauthor{\bsnm{Bovik}, \binits{A.C.}}:
\bctitle{Multiscale structural similarity for image quality assessment},
vol. \bseriesno{2},
pp. \bfpage{1398}--\blpage{14022}
(\byear{2003}).
\doiurl{10.1109/ACSSC.2003.1292216}
\end{bchapter}
\endbibitem

\bibitem[\protect\citeauthoryear{Wallace}{1975}]{wallace1975diurnal}
\begin{barticle}
\bauthor{\bsnm{Wallace}, \binits{J.M.}}:
\batitle{Diurnal variations in precipitation and thunderstorm frequency over
  the conterminous united states}.
\bjtitle{Monthly Weather Review}
\bvolume{103}(\bissue{5}),
\bfpage{406}--\blpage{419}
(\byear{1975})
\end{barticle}
\endbibitem

\bibitem[\protect\citeauthoryear{Joshi et~al.}{2021}]{joshi2021review}
\begin{barticle}
\bauthor{\bsnm{Joshi}, \binits{G.}},
\bauthor{\bsnm{Walambe}, \binits{R.}},
\bauthor{\bsnm{Kotecha}, \binits{K.}}:
\batitle{A review on explainability in multimodal deep neural nets}.
\bjtitle{IEEE Access}
\bvolume{9},
\bfpage{59800}--\blpage{59821}
(\byear{2021})
\end{barticle}
\endbibitem

\bibitem[\protect\citeauthoryear{Price et~al.}{2024}]{price_probabilistic_2024}
\begin{barticle}
\bauthor{\bsnm{Price}, \binits{I.}},
\bauthor{\bsnm{Sanchez-Gonzalez}, \binits{A.}},
\bauthor{\bsnm{Alet}, \binits{F.}},
\bauthor{\bsnm{Andersson}, \binits{T.R.}},
\bauthor{\bsnm{El-Kadi}, \binits{A.}},
\bauthor{\bsnm{Masters}, \binits{D.}},
\bauthor{\bsnm{Ewalds}, \binits{T.}},
\bauthor{\bsnm{Stott}, \binits{J.}},
\bauthor{\bsnm{Mohamed}, \binits{S.}},
\bauthor{\bsnm{Battaglia}, \binits{P.}},
\bauthor{\bsnm{Lam}, \binits{R.}},
\bauthor{\bsnm{Willson}, \binits{M.}}:
\batitle{Probabilistic weather forecasting with machine learning}.
\bjtitle{Nature}
(\byear{2024})
\doiurl{10.1038/s41586-024-08252-9} .
Accessed 2024-12-16
\end{barticle}
\endbibitem

\bibitem[\protect\citeauthoryear{Xia et~al.}{2024}]{xia2024accurate}
\begin{barticle}
\bauthor{\bsnm{Xia}, \binits{P.}},
\bauthor{\bsnm{Zhang}, \binits{L.}},
\bauthor{\bsnm{Min}, \binits{M.}},
\bauthor{\bsnm{Li}, \binits{J.}},
\bauthor{\bsnm{Wang}, \binits{Y.}},
\bauthor{\bsnm{Yu}, \binits{Y.}},
\bauthor{\bsnm{Jia}, \binits{S.}}:
\batitle{Accurate nowcasting of cloud cover at solar photovoltaic plants using
  geostationary satellite images}.
\bjtitle{Nature Communications}
\bvolume{15}(\bissue{1}),
\bfpage{510}
(\byear{2024})
\end{barticle}
\endbibitem

\bibitem[\protect\citeauthoryear{Nichol and
  Dhariwal}{2021}]{nichol2021improved}
\begin{bchapter}
\bauthor{\bsnm{Nichol}, \binits{A.Q.}},
\bauthor{\bsnm{Dhariwal}, \binits{P.}}:
\bctitle{Improved denoising diffusion probabilistic models}.
In: \bbtitle{International Conference on Machine Learning},
pp. \bfpage{8162}--\blpage{8171}
(\byear{2021}).
\bcomment{PMLR}
\end{bchapter}
\endbibitem

\bibitem[\protect\citeauthoryear{Voleti et~al.}{2022}]{voleti2022mcvd}
\begin{barticle}
\bauthor{\bsnm{Voleti}, \binits{V.}},
\bauthor{\bsnm{Jolicoeur-Martineau}, \binits{A.}},
\bauthor{\bsnm{Pal}, \binits{C.}}:
\batitle{Mcvd-masked conditional video diffusion for prediction, generation,
  and interpolation}.
\bjtitle{Advances in neural information processing systems}
\bvolume{35},
\bfpage{23371}--\blpage{23385}
(\byear{2022})
\end{barticle}
\endbibitem

\bibitem[\protect\citeauthoryear{Paszke et~al.}{2017}]{Paszke2017}
\begin{bchapter}
\bauthor{\bsnm{Paszke}, \binits{A.}}, \betal:
\bctitle{Automatic differentiation in pytorch}.
In: \bbtitle{NIPS 2017 Workshop on Autodiff}
(\byear{2017})
\end{bchapter}
\endbibitem

\bibitem[\protect\citeauthoryear{Loshchilov}{2017}]{loshchilov2017decoupled}
\begin{botherref}
\oauthor{\bsnm{Loshchilov}, \binits{I.}}:
Decoupled weight decay regularization.
arXiv preprint arXiv:1711.05101
(2017)
\end{botherref}
\endbibitem

\bibitem[\protect\citeauthoryear{Morales-Brotons
  et~al.}{2024}]{morales2024exponential}
\begin{botherref}
\oauthor{\bsnm{Morales-Brotons}, \binits{D.}},
\oauthor{\bsnm{Vogels}, \binits{T.}},
\oauthor{\bsnm{Hendrikx}, \binits{H.}}:
Exponential moving average of weights in deep learning: Dynamics and benefits.
Transactions on Machine Learning Research
(2024)
\end{botherref}
\endbibitem

\bibitem[\protect\citeauthoryear{Ho and Salimans}{2022}]{ho2022classifier}
\begin{botherref}
\oauthor{\bsnm{Ho}, \binits{J.}},
\oauthor{\bsnm{Salimans}, \binits{T.}}:
Classifier-free diffusion guidance.
arXiv preprint arXiv:2207.12598
(2022)
\end{botherref}
\endbibitem

\bibitem[\protect\citeauthoryear{Lin
  et~al.}{2024}]{lin2024commondiffusionnoiseschedules}
\begin{botherref}
\oauthor{\bsnm{Lin}, \binits{S.}},
\oauthor{\bsnm{Liu}, \binits{B.}},
\oauthor{\bsnm{Li}, \binits{J.}},
\oauthor{\bsnm{Yang}, \binits{X.}}:
Common Diffusion Noise Schedules and Sample Steps are Flawed
(2024).
\url{https://arxiv.org/abs/2305.08891}
\end{botherref}
\endbibitem

\bibitem[\protect\citeauthoryear{Rasp et~al.}{2020}]{Rasp2020WeatherBenchAB}
\begin{botherref}
\oauthor{\bsnm{Rasp}, \binits{S.}},
\oauthor{\bsnm{Dueben}, \binits{P.D.}},
\oauthor{\bsnm{Scher}, \binits{S.}},
\oauthor{\bsnm{Weyn}, \binits{J.A.}},
\oauthor{\bsnm{Mouatadid}, \binits{S.}},
\oauthor{\bsnm{Thuerey}, \binits{N.}}:
Weatherbench: A benchmark data set for data‐driven weather forecasting.
Journal of Advances in Modeling Earth Systems
\textbf{12}
(2020)
\end{botherref}
\endbibitem

\bibitem[\protect\citeauthoryear{Chen et~al.}{2025}]{chen_2025_14643154}
\begin{botherref}
\oauthor{\bsnm{Chen}, \binits{H.}},
\oauthor{\bsnm{Zhong}, \binits{X.}},
\oauthor{\bsnm{Zhai}, \binits{Q.}},
\oauthor{\bsnm{Li}, \binits{X.}},
\oauthor{\bsnm{Chan}, \binits{Y.W.}},
\oauthor{\bsnm{Chan}, \binits{P.W.}},
\oauthor{\bsnm{Huang}, \binits{Y.}},
\oauthor{\bsnm{Li}, \binits{H.}},
\oauthor{\bsnm{Shi}, \binits{X.}}:
Skillful Nowcasting of Convective Clouds with a Cascade Diffusion Model.
Zenodo
(2025).
\doiurl{10.5281/zenodo.14643154} .
\url{https://doi.org/10.5281/zenodo.14643154}
\end{botherref}
\endbibitem

\end{thebibliography}

\end{document}